\pdfoutput=1

\documentclass[11pt]{article}

\usepackage[final]{acl}

\usepackage{times}
\usepackage{latexsym}
\usepackage{tabularx}
\usepackage{amsmath, amssymb, amsfonts}
\usepackage{algorithm}
\usepackage{algorithmicx}
\usepackage{algpseudocode}
\usepackage{multirow}
\usepackage{booktabs}
\usepackage{float}
\usepackage{placeins}
\usepackage[T1]{fontenc}

\usepackage[utf8]{inputenc}

\usepackage{microtype}

\usepackage{inconsolata}
\usepackage{subcaption}
\usepackage{graphicx}

\title{SPFT-SQL: Enhancing Large Language Model for Text-to-SQL Parsing by Self-Play Fine-Tuning}

\author{
 \textbf{Yuhao Zhang\textsuperscript{1,3}},
 \textbf{Shaoming Duan\textsuperscript{\dag2}},
 \textbf{Jinhang Su\textsuperscript{1}},
 \textbf{Chuanyi Liu\textsuperscript{\dag1,2}},
\\
 \textbf{Peiyi Han\textsuperscript{1,2}},
\\
\\
 \textsuperscript{1}Harbin Institute of Technology, Shenzhen,
 \textsuperscript{2}Pengcheng Laboratory,\\
 \textsuperscript{3}Mindflow.ai, \\
\\
 \small{
   \textbf{Correspondence:} \href{mailto:email@domain}
   {shaomingduan@gmail.com}, {liuchuanyi@hit.edu.cn}
 }
}

\begin{document}
\renewcommand{\thefootnote}{}
\maketitle
\begin{abstract}

Despite the significant advancements of self-play fine-tuning (SPIN), which can transform a weak large language model (LLM) into a strong one through competitive interactions between models of varying capabilities, it still faces challenges in the Text-to-SQL task. SPIN does not generate new information, and the large number of correct SQL queries produced by the opponent model during self-play reduces the main model's ability to generate accurate SQL queries.  To address this challenge, we propose a new self-play fine-tuning method tailored for the Text-to-SQL task, called SPFT-SQL. Prior to self-play, we introduce a verification-based iterative fine-tuning approach, which synthesizes high-quality fine-tuning data iteratively based on the database schema and validation feedback to enhance model performance, while building a model base with varying capabilities. During the self-play fine-tuning phase, we propose an error-driven loss method that incentivizes incorrect outputs from the opponent model, enabling the main model to distinguish between correct SQL and erroneous SQL generated by the opponent model, thereby improving its ability to generate correct SQL.  Extensive experiments and in-depth analyses on six open-source LLMs and five widely used benchmarks demonstrate that our approach outperforms existing state-of-the-art (SOTA) methods.

\footnote{\dag ~Corresponding authors}

\end{abstract}

\section{Introduction}
Text-to-SQL \cite{qin2022survey, li2024can} aims to automatically convert natural language questions into SQL queries, enabling non-expert users to easily retrieve information from databases. Recent studies \cite{sun2024sqlpalm, li2024codes, dtssq12024} have demonstrated that supervised fine-tuning (SFT) \cite{ouyang2022sft} can significantly enhance performance on Text-to-SQL tasks by transforming a general-purpose open-source LLM into a specialized one. Additionally, SFT-based approaches have gained widespread research attention due to their potential to address privacy risks and reduce overhead associated with closed-source LLMs (e.g., GPT-4 \cite{achiam2023gpt}) \cite{gao2023text, pourreza2024din, lee2024mcs}. However, a major challenge for SFT-based methods is the high cost of acquiring Text-to-SQL data, which typically requires manual expert annotation.


\begin{figure}[t]
    \centering
    \includegraphics[width=\linewidth]{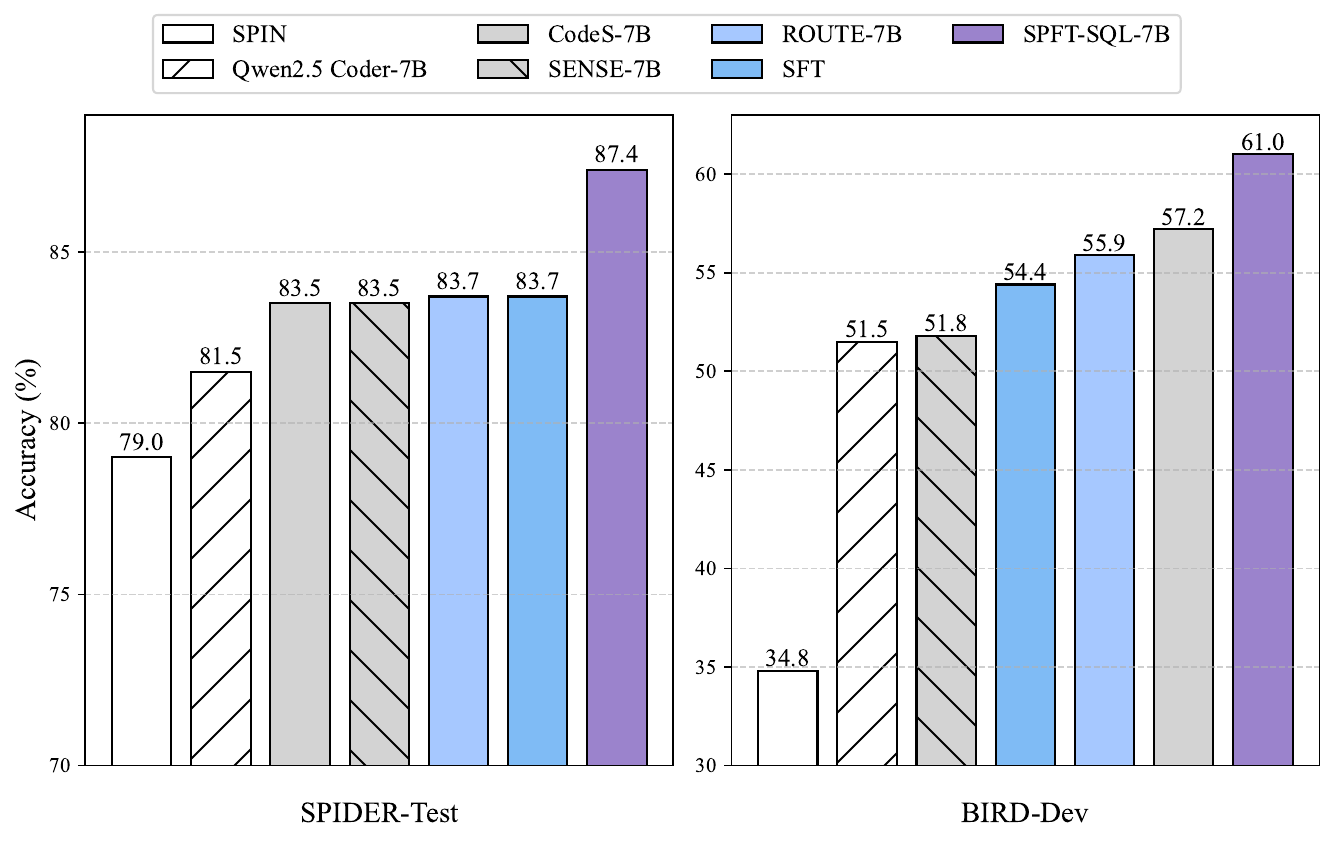}
    \caption{Comparison results on the Spider \cite{yu2018spider} and BIRD\cite{li2024can} dataset, the base model of SFT, SPIN \cite{chen2024selfplay}, and SPFT-SQL is Qwen2.5-Coder 7B.}
    \label{fig:evaluation_results}
\end{figure}


To address this issue, recent efforts \cite{yang2024synthesizing, li2024codes, zhang2024sciencebenchmark} have proposed data synthesis strategies for generating Text-to-SQL data and fine-tuning open-source models, yielding significant performance improvements (see Figure \ref{fig:evaluation_results}). However, these methods still rely on closed-source LLMs, such as GPT-3.5/4 \cite{achiam2023gpt}, for data synthesis, raising privacy concerns. In response, ROUTE \cite{qin2025route} introduced a method for synthesizing fine-tuning data for tasks like Text-to-SQL and Schema Linking using open-source models, improving model generalization through multi-task supervised fine-tuning and achieving a new state-of-the-art (SOTA) performance. However, the limited generation capacity of open-source models restricts the quality of synthetic data, which in turn limits model performance.


An alternative approach involves iteratively synthesizing data through self-play fine-tuning (SPIN) \cite{chen2024selfplay, cheng2025self, wu2024self} to transform a weak LLM into a stronger one. Self-play, which has been successfully applied in domains such as reasoning \cite{cheng2025self}, AlphaGo \cite{AlphaGo}, and AlphaZero \cite{AlphaZero}, enables models to compete with themselves at various stages, enhancing both performance and data synthesis capabilities while overcoming the limitations of open-source model generation. In the context of the Text-to-SQL task, the only prior work \cite{liu2022augmenting} applied self-play to multi-turn Text-to-SQL, generating multiple rounds of intermediate questions and answers for data augmentation. While this method improved performance in multi-turn tasks, it is not applicable to single-turn Text-to-SQL, as it only generates intermediate data based on existing annotated pairs.


This motivates us to conduct a thorough evaluation of SPIN in the Text-to-SQL task, assessing its potential as an alternative approach. As shown in Figure \ref{fig:evaluation_results}, applying the existing SPIN method \cite{chen2024selfplay} to Text-to-SQL results in a significant performance drop, which is much lower than that of SFT-based methods utilizing existing data synthesis techniques \cite{qin2025route, li2024codes, yang2024synthesizing}. A subsequent analysis of failure cases reveals two key challenges for SPIN in the Text-to-SQL domain. First, SPIN only synthesizes SQL queries from existing natural language questions, without generating new information. This limitation restricts the model's ability to improve, and repeated training leads to overfitting. Second, the self-play mechanism in SPIN treats all data generated by the opponent model as incorrect, which results in many valid SQL queries being discarded as erroneous, thus hindering the model's ability to learn from errors.

To address these challenges, we propose a self-play fine-tuning method for Text-to-SQL tasks, called SPFT-SQL. Specifically, prior to self-play, we introduce a verification-based iterative supervised fine-tuning approach that iteratively synthesizes high-quality data for fine-tuning the LLM. This method randomly selects schemas (e.g., tables and columns) from the database and combines them with SQL templates to generate executable SQL queries. Corresponding natural language questions (NLQs) are then synthesized using a SQL-to-Text model. The synthesized NLQ-SQL pairs are used to fine-tune the Text-to-SQL model, enhancing its performance. The SQL-to-Text model is subsequently updated with the synthesized data that passes a verification strategy. During self-play fine-tuning, the strongest model from the previous stage serves as the main model, while the weakest model acts as the opponent. We introduce an error-driven loss function that penalizes correct SQL queries generated by the opponent model and incentivizes the generation of incorrect queries. This mechanism enables the main model to better distinguish between correct and incorrect results, thus improving its ability to generate correct SQL queries. In the next iteration, the newly acquired main model is incorporated into the next round of supervised fine-tuning.



\begin{figure*}[ht]
  \includegraphics[width=\textwidth]{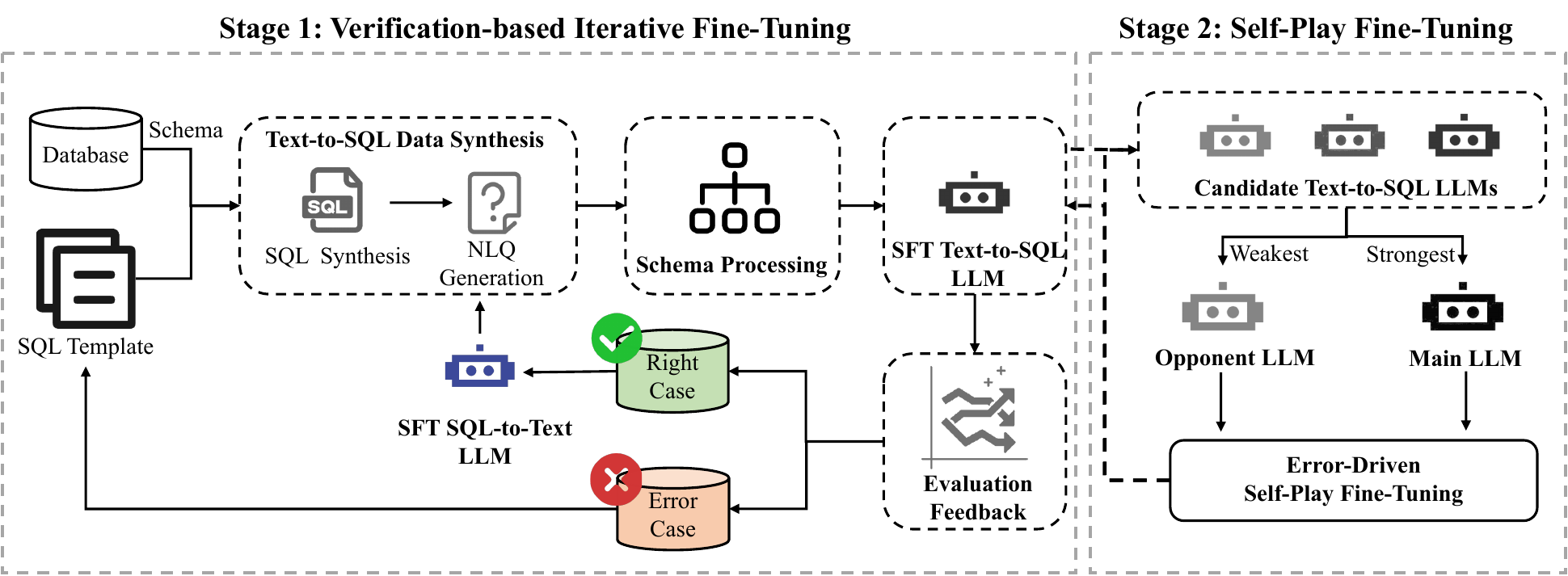}
  \caption {An overview of SPFT-SQL framework.}
  \label{fig:SPFT-SQL}
\end{figure*}

The main contributions of this work are as follows:
\begin{itemize}
    \item We first evaluated the performance of the SPIN method on the Text-to-SQL task and found that the existing SPIN method performs poorly in this context. This prompted us to propose a new self-play fine-tuning method specifically designed for the Text-to-SQL task.
    \item We propose a verification-based iterative fine-tuning framework that synthesizes data iteratively based on the database schema and improves data quality through verification feedback, thereby continuously enhancing model performance.
    \item We introduce an error-driven loss that penalizes the generation of incorrect outputs by the opponent model during the self-play fine-tuning phase. This enables the main model to distinguish between correct SQL and erroneous SQL generated by the opponent model, ultimately improving the main model's ability to generate accurate SQL queries.
    \item Extensive experiments on five datasets and six open-source LLMs of varying types and parameter sizes. The results demonstrate that our approach not only effectively improves model performance but also outperforms other SOTA methods based on open-source models. Furthermore, after fine-tuning with our method, small-parameter open-source models outperform methods based on large-parameter, closed-source LLM.
    

\end{itemize}

\section{Related Works}

\textbf{Self-Play Fine-Tuning} \quad Self-play \cite{zhang2024survey, digiovanni2021survey}, where the algorithm learns by competing against itself, has gained significant attention due to its success in AlphaGo \cite{AlphaGo} and AlphaZero \cite{AlphaZero}. To transform a weak LLM into a stronger one, existing studies \cite{chen2024selfplay, alami2024investigating, yin2024self, wu2025selfplay} have proposed introducing self-play mechanisms into LLMs without requiring additional human-annotated data. In the text-to-SQL task, there is only one prior work \cite{liu2022augmenting} that applies self-play to text-to-SQL. However, this method only uses self-play to generate multiple rounds of intermediate data based on existing annotated data, which makes it inapplicable to single-turn text-to-SQL tasks. In contrast to previous studies, our SPFT-SQL introduces self-play fine-tuning into the text-to-SQL task by iteratively synthesizing new text-to-SQL pairs for data augmentation. Furthermore, we propose an error-incentive loss that encourages the generation of erroneous outputs by the opponent model, thereby enhancing the main model's ability to generate correct SQL queries.

\paragraph{SFT-based Text-to-SQL}  To improve the performance of open-source LLMs on text-to-SQL tasks, existing research \cite{sun2024sqlpalm, chen2024open, dtssq12024} has applied supervised fine-tuning on annotated data. However, a key challenge remains the high cost of human-annotated data. To reduce this cost, some efforts \cite{li2024codes, yang2024synthesizing} have employed various data synthesis strategies, using LLMs to generate data for fine-tuning. However, these methods rely on the general capabilities of closed-source LLMs, such as GPT-4, which raises privacy concerns. To address this issue, Route \cite{qin2025route} proposed a data augmentation approach to improve generalization using open-source LLMs. In contrast to previous work, our SPFT-SQL method iteratively synthesizes high-quality data through the self-play mechanism.




\section{Methodology}

The SPFT-SQL framework consists of two stages: Verification-Based Iterative Fine-Tuning and Self-Play Fine-Tuning, as illustrated in Figure \ref{fig:SPFT-SQL}. In the first stage, verification-based iterative fine-tuning continuously generates high-quality data for fine-tuning, producing various candidate text-to-SQL models for the subsequent self-play phase. During self-play fine-tuning, the strongest model from the previous stage serves as the main model, while the weakest model functions as the opponent. Using the proposed error-driven loss, self-play fine-tuning is applied between the opponent model and the main model to enhance the main model's ability to generate correct SQL queries from a NLQ. This process is iterated until the model converges.





\subsection{Verification-Based Iterative Fine-Tuning}

Verification-based iterative fine-tuning generates high-quality data and diverse candidate models for self-play. As shown in Figure~\ref{fig:SPFT-SQL}, synthetic data fine-tunes the Text-to-SQL model, which then verifies validation data. Verified samples improve the SQL-to-Text model, while failed cases provide templates for the next iteration, jointly enhancing data quality and model performance.

\subsubsection{Text-to-SQL Data Synthesis}

The Text-to-SQL fine-tuning data synthesis process begins by generating the SQL query, followed by the synthesis of the corresponding NLQ.


\paragraph{SQL Synthesis}

To generate new SQL queries based on training data schemas, we employ a template-based approach as outlined by \cite{hu2023importance}. First, a pool of SQL templates is created by normalizing schema-related mentions (columns and values) and removing JOIN phrases. A template is then sampled according to the training distribution, and tables and columns are selected with constraints to fill the normalized slots within the template.

To accurately extract the SQL template while preserving key relationships, we leverage the general understanding capabilities of the LLM. The prompt used for this extraction is defined in Appendix \ref{sec:Template_Extraction_prompt}. The final template maintains the query structure and data types, allowing it to adapt to various query scenarios. By omitting the \texttt{FROM} and \texttt{JOIN} clauses, the template becomes independent of specific table names, yet it retains essential query structures (e.g., \texttt{SELECT}, \texttt{WHERE}, \texttt{GROUP BY}, \texttt{HAVING}) to ensure consistency. Foreign key relationships are denoted using a special format (e.g., \texttt{col\_number\_key\_fk}).

Once the template is generated, the method takes as input the database \( d \) and the SQL template \( t = (q, c, v) \), where \( t \) consists of the query structure \( q \), the set of columns \( c \), and the set of values \( v \). For columns \( c_1 \) to \( c_m \) in the template, a column is randomly selected and replaced from those that match the data types in the database. During the column selection process, if a column \( z \) comes from an already selected table, it is assigned a weight \( p = 1 \); otherwise, the weight is adjusted based on schema distance and accumulated through iterations, ensuring that the final column selection adheres to database schema consistency. After filling in the columns, corresponding values \( v_1 \) to \( v_n \) are retrieved and randomly filled from the database. This process leverages the database schema information to ensure that both column selection and value filling respect logical constraints and data type matching, thereby generating structurally consistent and logically sound SQL statements.

\paragraph{NLQ Synthesis}

To ensure that the synthesized NLQ aligns with the intent of the SQL query, we iteratively fine-tune a SQL-to-Text model to generate the corresponding NLQ based on a given SQL query. The fine-tuning  SQL-NLQ pairs are derived from the correct data synthesized in the previous iteration. As self-play progresses, the model's performance improves, leading to higher-quality synthetic data and, in turn, enhanced performance of the SQL-to-Text model.



\subsubsection{Schema Processing and SFT}

To effectively utilize the Text-to-SQL synthetic data for fine-tuning LLMs, we address the challenge of capturing implicit patterns between database schemas and NLQs, which is complicated by the complexity of database structures. Inspired by \cite{li2024codes}, we introduce schema processing during SFT, employing three strategies: Structured Schema Extraction, Context-Aware Value Matching, and Database Metadata Augmentation.

Structured Schema Extraction filters irrelevant information by selecting the most relevant tables and columns, improving the model's focus on the database structure. Context-Aware Value Matching enhances the query-database association by aligning query columns with their corresponding values, ensuring more accurate SQL conditions. Finally, Database Metadata Augmentation incorporates metadata such as key relationships, data types, and annotations, providing richer context for understanding table relationships and field semantics. These strategies work together to progressively enhance the model’s SQL generation capabilities.


\subsubsection{Evaluation Feedback}


The fine-tuned Text-to-SQL model is evaluated on the synthesized validation set \(D_{\text{val}} \) using an SQL executor $E$, and samples are classified based on execution results:

\begin{equation}
\left\{
\begin{array}{ll}
    y_+ = y', & \text{if } \mathbf{E}(y') = \mathbf{E}(y) \\[4pt]
    y_- = y', & \text{if } \mathbf{E}(y') \neq \mathbf{E}(y)
\end{array}
\right.
\end{equation}


where \( y' \) represents the generated results of the fine-tuned Text-to-SQL model on \(D_{\text{val}} \), while \( y \) refers to the ground truth SQL in \(D_{\text{val}} \). To enhance the generalization ability of SQL-to-Text, $y_+$ is used as training data for the next iteration. This serves as positive feedback, boosting the model's ability to generate diverse question-answer pairs. For incorrect samples (\(y_-\)), the SQL templates from these queries are selected for the next iteration to generate new SQL-question pairs, allowing the model to correct errors. By combining these two strategies, a collaborative optimization mechanism is established, progressively reducing the proportion of incorrect samples while improving the quality of the training data. As the iterative fine-tuning continues, both the quantity and diversity of synthetic data increase, leading to improved model performance. 

\subsection{Self-Play Fine-Tuning}


Based on the evaluation accuracy on the synthetic validation data \( D_{\text{val}} \), the model with the highest accuracy is selected as the main model \( p_{\theta_{m}} \), and the model with the lowest accuracy is chosen as the opponent model \( p_{\theta_{o}} \). The self-play fine-tuning aims to train the main model to distinguish between correct and incorrect SQL generated by the opponent, thereby enhancing its ability to generate accurate SQL, as outlined in Algorithm \ref{alg:self-play}. To achieve this, we propose an error-driven loss function that penalizes incorrect SQL generated by the opponent, using a defined reward signal: 
\begin{align}
\noindent
\label{eq:reward_signal}
\small
    R(y', x) = 
    \begin{cases}
    1, & \text{if } \mathbf{E}(y') \neq \mathbf{E}(y) \\[4pt]
    0, & \text{if } \mathbf{E}(y') = \mathbf{E}(y)
    \end{cases}
\end{align}

Based on this reward signal, we formulate the error-driven loss as:
\begin{align}&
\noindent 
\label{eq:self_play_loss}
\small
    \mathcal{L}_{\text{Self-Play}} = 
    \mathbb{E} 
    \Bigg[ \ell \small \Bigg( \lambda R(y',x)*\log \frac{p_{\theta_{m}} (y_+ | x)}{p_{\theta_{o}} (y_+ | x)}  \notag \\ &\quad
    - \lambda R(y',x)*\log \frac{p_{\theta_{m}} (y_- | x)}{p_{\theta_{o}} (y_- | x)} \Bigg) \Bigg]
\end{align}
where \( x \) is the natural language question, \( \lambda \) is the regularization parameter, and \( \ell \) represents a convex and decreasing loss function, for which we adopt the logistic loss function following SPIN.

\begin{algorithm}
\noindent
\caption{Self-Play Fine-Tuning}
\label{alg:self-play}
\begin{algorithmic}
\State \textbf{Input:} Candidate model set $\mathcal{M} = \{p_{\theta_0}, p_{\theta_1}, ..., p_{\theta_n}\}$, validation dataset $\mathcal{D}_{val}$, preference scaling $\beta$, max iterations $T$ 
\State \textbf{Output:} Optimized model $p_{\theta_{m}}$ 
\For{$t = 1$ to $T$}  
    \State \textbf{Model Selection:}  
    \State $p_{\theta_{m}} = \arg\max_{p_{\theta} \in \mathcal{M}} \text{Acc}(p_{\theta}, \mathcal{D}_{val})$
    \State $p_{\theta_{o}} = \arg\min_{p_{\theta} \in \mathcal{M}} \text{Acc}(p_{\theta}, \mathcal{D}_{val})$

    \State \textbf{Preference Data Generation:}  
    \State Use $p_{\theta_{o}}$ to generate pairs $(y_+, y_-)$ on $\mathcal{D}_{val}$  

    \State \textbf{Model Optimization:}  
    \State Update $p_{\theta_{m}}$ using Equation (\ref{eq:self_play_loss})
    
    \State Add optimized $p_{\theta_{m}}$ to $\mathcal{M}$

\EndFor  
\State \textbf{Return:}  $p_{\theta_{m}}$ 
\noindent
\end{algorithmic}  
\end{algorithm}

This loss function reinforces the main model’s ability to generate correct SQL by comparing the weighted probability differences between \(p_{\theta_{m}}\) and \(p_{\theta_{o}}\) for correct SQL (\(y_{+}\)) and incorrect SQL (\(y_{-}\)), modulated by the reward signal \(R(y',x)\) and regularization parameter \(\lambda\). The first term encourages the main model to assign higher probability to correct SQL than the opponent model. The second term penalizes the main model for behaving similarly to the opponent model on incorrect SQL. This approach enables the main model to learn and avoid the error patterns exhibited by the opponent model in SQL generation.

Compared with Direct Preference Optimization (DPO)~\cite{rafailov2024direct}, which defines an implicit reward $r^*(x,y) = \beta \log\frac{\pi_\theta(y|x)}{\pi_{\text{ref}}(y|x)}$ based on the Bradley-Terry model, our method provides clearer and more actionable optimization signals. DPO estimates the relative advantage of a policy over a fixed reference model, but its reward vanishes when their outputs are similar—a common scenario in Text-to-SQL tasks due to semantically equivalent SQL queries—thus limiting learning. In contrast, our execution-based reward $R(y', x)$ directly penalizes incorrect outputs, encouraging effective error correction. Additionally, DPO adopts a single-step training paradigm with a static reference, restricting its adaptability. Our method introduces an iterative self-play strategy with a dynamically updated opponent, continuously increasing the training challenge. This progressive optimization drives the main policy toward the optimal distribution, mitigating local optima and enhancing robustness. More details refer to Appendix~\ref{Appendix:Dpo_Analysis} and~\ref{Appendix:dpo_ppo_ablation}.

\section{Experiments}

\begin{table*}[ht!]
\centering
\fontsize{9}{10.2}\selectfont
\setlength{\tabcolsep}{5pt}
\setlength{\abovecaptionskip}{2pt}
\begin{tabular}{lccccc|cccccc}
\toprule
& \multicolumn{3}{c}{\textbf{SPIDER}} & \multicolumn{2}{c|}{\textbf{BIRD}} & \multicolumn{5}{c}{\textbf{SPIDER-Variants}}\\
\cmidrule(lr){2-4} \cmidrule(lr){5-6} \cmidrule(lr){7-11}
\textbf{Methods} & \multicolumn{2}{c}{\textbf{Dev}} & \textbf{Test} & \multicolumn{2}{c|}{\textbf{Dev}} & \multicolumn{2}{c}{\textbf{Syn}} & \multicolumn{2}{c}{\textbf{Realistic}} & \textbf{DK} \\
\cmidrule(lr){2-3}  \cmidrule(lr){4-4} \cmidrule(lr){5-6} \cmidrule(lr){7-8} \cmidrule(lr){9-10}  \cmidrule(lr){11-11}
 & \textbf{EX} & \textbf{TS} & \textbf{EX} & \textbf{EX} & \textbf{VES} & \textbf{EX} & \textbf{TS} & \textbf{EX} & \textbf{TS} & \textbf{EX} \\
\midrule
\multicolumn{11}{l}{\textit{Prompting Closed-Source LLMs(As a reference)}} \\
GPT-4 \cite{achiam2023gpt} & 72.9 & 64.9 & 76.1 & 46.4 & 49.8 & 64.0 & 54.7 & 65.7 & 54.9 & 59.3 \\
DIN-SQL + GPT4 \cite{pourreza2024din} & 82.8 & 74.2 & 85.3 & 50.7 & 58.8 & 68.3 & 61.9 & 71.3 & 64.8 & 66.7 \\
MAC-SQL + GPT4 \cite{wang2025mac} & 86.8 & - & 82.8 & 59.4 & 66.2 & 72.5 & 61.5 & 79.9 & 65.4 & 71.4 \\
DAIL-SQL + GPT4 \cite{gao2023text} & 83.5 & 76.2 & 86.6 & 54.8 & 56.1 & 68.7 & 60.7 & 77.2 & 68.5 & 66.5 \\
MCS-SQL + GPT4 \cite{lee2024mcs} & 89.5 & - & 89.6 & 63.4 & 64.8 & - & - & - & - & - \\

\midrule
\multicolumn{11}{l}{\textit{Fine-Tuning Open-Source LLMs(1.5B)}} \\
Qwen2.5 Coder-1.5B \cite{hui2024qwen2} & 72.4 & 62.5 & 72.3 & 40.6 & 43.3 & 55.7 & 45.4 & 60.4 & 46.5 & 62.0 \\
SFT + Qwen2.5 Coder-1.5B \cite{hui2024qwen2} & 76.8 & 70.2 & 78.0 & 44.3 & 45.6 & 59.0 & 51.6 & 64.8 & 56.3 & 60.9 \\
SFT + CodeS-1B \cite{li2024codes} & 77.9 & 72.2 & - & 50.4 & 51.0 & 66.5 & 59.3 & 70.9 & 61.8 & 64.7 \\
SPIN + Qwen2.5 Coder-1.5B \cite{chen2024selfplay} & 67.6 & 60.5 & 68.5 & 21.0 & 22.1 & 55.0 & 46.4 & 57.2 & 45.5 & 54.0 \\
\textbf{SPFT-SQL + Qwen2.5 Coder-1.5B} & \textbf{79.7} & \textbf{73.5} & \textbf{82.3} & \textbf{54.0} & \textbf{59.9} & \textbf{66.7} & \textbf{59.4} & \textbf{75.4} & \textbf{67.3} & \textbf{67.3} \\

\midrule
\multicolumn{11}{l}{\textit{Fine-Tuning Open-Source LLMs(7B)}} \\
Llama3-8B \cite{touvron2023llama} & 72.3 & 63.9 & 69.6 & 39.2 & 43.3 & 60.3 & 51.2 & 62.0 & 50.4 & 57.4 \\
Deepseek-7B \cite{guo2024Deepseek} & 67.0 & 57.7 & 69.4 & 40.1 & 44.5 & 55.3 & 46.0 & 57.7 & 45.9 & 55.3 \\
Qwen2.5 Coder-7B \cite{hui2024qwen2} & 83.5 & 79.2 & 81.5 & 51.5 & 55.3 & 69.8 & 64.2 & 75.4 & 70.9 & 68.0 \\
SFT + Llama3-8B \cite{touvron2023llama} & 79.5 & 73.6 & 80.9 & 51.8 & 55.3 & 66.4 & 60.1 & 71.1 & 62.8 & 61.7 \\
SFT + Deepseek-7B \cite{guo2024Deepseek} & 78.6 & 71.9 & 81.5 & 53.9 & 57.1 & 64.8 & 57.5 & 69.3 & 61.4 & 60.7 \\
SFT + Qwen2.5 Coder-7B \cite{hui2024qwen2} & 82.9 & 79.0 & 83.7 & 54.4 & 56.1 & 68.3 & 62.3 & 75.2 & 69.5 & 66.5 \\
SENSE-7B \cite{yang2024synthesizing} & 83.2 & \textbf{81.7} & 83.5 & 51.8 & - & 72.6 & 64.9 & 82.7 & 75.6 & \textbf{77.9} \\
DTS-SQL-7B \cite{dtssq12024} & 85.5 & - & 84.4 & 55.8 & 60.3 & - & - & - & - & - \\
SFT + CodeS-7B \cite{li2024codes} & 85.4 & 80.3 & 83.5 & 57.2 & 58.8 & \underline{76.9} & 70.0 & 82.9 & \underline{77.2} & 72.0 \\
ROUTE + Llama3-8B \cite{qin2025route} & \underline{86.0} & 80.3 & 83.9 & 57.3 & 60.1 & \textbf{77.4} & \textbf{70.2} & 80.9 & 72.6 & 74.6 \\
ROUTE + Qwen2.5-7B \cite{qin2025route} & 83.6 & 77.5 & 83.7 & 55.9 & 57.4 & - & - & - & - & - \\
SPIN + Deepseek-7B \cite{chen2024selfplay} & 61.8 & 51.5 & 63.9 & 27.0 & 29.1 & 45.6 & 35.8 & 52.2 & 37.6 & 48.2 \\
SPIN + Llama3-8B \cite{chen2024selfplay} & 79.8 & 73.6 & 80.2 & 32.4 & 37.3 & 66.8 & 60.2 & 69.3 & 60.0 & 61.9 \\
SPIN + Qwen2.5 Coder-7B \cite{chen2024selfplay} & 78.0 & 73.1 & 79.0 & 34.8 & 37.7 & 71.2 & 57.5 & 63.6 & 65.9 & 61.8 \\
\textbf{SPFT-SQL + Llama3-8B} & 83.0 & 75.4 & \underline{86.4} & 60.6 & 65.4 & 76.1 & 69.1 & \textbf{85.0} & \textbf{77.6} & 74.2 \\
\textbf{SPFT-SQL + Deepseek-7B} & 82.3 & 78.0 & 86.0 & 58.3 & 64.0 & 76.5 & \underline{70.0} & 80.3 & 74.8 & 72.5 \\
\textbf{SPFT-SQL + Qwen2.5 Coder-7B} & \textbf{87.2} & \underline{81.3} & \textbf{87.4} & \textbf{61.0} & \textbf{67.0} & 75.1 & 67.6 & \underline{83.3} & 75.6 & \underline{75.5} \\

\midrule
\multicolumn{11}{l}{\textit{Fine-Tuning Open-Source LLMs(14B)}} \\
Qwen2.5 Coder-14B \cite{hui2024qwen2} & 83.8 & 78.0 & 84.9 & 58.0 & 62.8 & 74.3 & 66.8 & 76.4 & 69.1 & 69.7 \\
SFT + Qwen2.5 Coder-14B \cite{hui2024qwen2} & 84.8 & 79.6 & 84.4 & 58.5 & 63.9 & 68.4 & 62.1 & 74.4 & 66.7 & 70.3 \\
SENSE-13B \cite{yang2024synthesizing} & 84.1 & \textbf{83.5} & 86.6 & 55.5 & - & \underline{77.6} & \underline{70.2} & \underline{84.1} & \underline{76.6} & \textbf{80.2} \\
SFT + CodeS-15B \cite{li2024codes} & 84.9 & 79.4 & 85.0 & 58.5 & 56.7 & 77.0 & 67.4 & 83.1 & 75.6 & 70.7 \\
ROUTE + Qwen2.5-14B \cite{qin2025route} & \underline{87.3} & 80.9 & \underline{87.1} & \underline{60.9} & \underline{65.2} & - & - & - & - & - \\
SPIN + Qwen2.5 Coder-14B \cite{chen2024selfplay} & 82.3 & 76.9 & 81.4 & 36.8 & 41.2 & 72.8 & 62.0 & 68.1 & 66.1 & 67.8 \\
\textbf{SPFT-SQL + Qwen2.5 Coder-14B} & \textbf{87.7} & \underline{81.9} & \textbf{89.0} & \textbf{63.6} & \textbf{68.9} & \textbf{78.4} & \textbf{71.3} & \textbf{84.6} & \textbf{77.6} & \underline{77.0} \\

\midrule
\multicolumn{11}{l}{\textit{Fine-Tuning Open-Source LLMs(32B)}} \\
Qwen2.5 Coder-32B \cite{hui2024qwen2} & 79.6 & 73.9 & 82.3 & 58.1 & 61.7 & 73.7 & 67.6 & 75.6 & 68.3 & 71.0 \\
SFT + Qwen2.5 Coder-32B \cite{hui2024qwen2} & 85.2 & 79.5 & 86.4 & 61.2 & 66.6 & 77.2 & 71.1 & 76.0 & 70.1 & 72.5 \\
OmniSQL-32B \cite{li2025omnisql} & 80.9 & - & {87.6} & {64.5} & - & 69.7 & - & 78.1 & - & \textbf{76.1} \\
SPIN + Qwen2.5 Coder-32B \cite{chen2024selfplay} & 79.2 & 72.8 & 80.8 & 40.1 & 44.3 & 72.9 & 64.2 & 67.5 & 64.1 & 69.3 \\
\textbf{SPFT-SQL + Qwen2.5 Coder-32B} & \textbf{87.8} & \textbf{83.2} & \textbf{89.1} & \textbf{65.2} & \textbf{70.5} & \textbf{81.7} & \textbf{72.3} & \textbf{86.2} & \textbf{76.8} & \underline{75.5} \\
\bottomrule
\end{tabular}
\caption{Performance of different methods on SPIDER, BIRD, and SPIDER-variants Datasets.}
\label{tab:combined_performance}
\end{table*}

\subsection{Experiment Setup}
\paragraph{Benchmarks}

To evaluate the effectiveness of our approach, we conduct experiments on five Text-to-SQL benchmarks, including the widely used cross-domain datasets SPIDER \cite{yu2018spider} and BIRD \cite{li2024can}, along with three SPIDER-derived versions: SPIDER-SYN \cite{gan2021towards}, SPIDER-Realistic \cite{deng2021realistic}, and SPIDER-DK \cite{gan2021dk}.SPIDER contains 7,000 training samples, 1,034 development samples, and 2,147 test samples, covering 206 databases across 138 domains.BIRD introduces more complex domain-specific queries, comprising 12,751 question-SQL pairs from 37 domains, including finance, healthcare, and education.SPIDER-SYN augments 20 SPIDER validation databases via synonym substitution,SPIDER-Realistic generates natural questions from 19 databases for realism, and SPIDER-DK introduces 535 knowledge-intensive queries over six databases to enhance domain reasoning.More experimental designs can be found in the Appendix \ref{Appendix:exp_design}.
\paragraph{Evaluation Metrics}
We evaluate model performance using execution accuracy (EX)\cite{yu2018spider} and test-suite accuracy (TS)\cite{zhong2020ts} on SPIDER and its variants. 
EX measures if the generated SQL matches the gold SQL execution, while TS verifies its performance across multiple test cases with database augmentation.
For BIRD, following its official settings, we report EX and Valid Efficiency Score (VES)\cite{li2024can}, which measures SQL execution efficiency. 

\paragraph{Models}
We evaluate the generalizability of our method using six open-source LLMs, including Llama3-8B-Instruct \cite{touvron2023llama}, Deepseek-Coder-7B-Instruct \cite{guo2024Deepseek}, and Qwen2.5-Coder \cite{hui2024qwen2} in sizes 1.5B, 7B, 14B, and 32B.

\paragraph{Baselines} We compare SPFT-SQL with a broad range of baselines, including both closed-source and open-source methods.Closed-source baselines include direct prompting with GPT-4~\cite{achiam2023gpt}, as well as enhanced prompting techniques such as DIN-SQL~\cite{pourreza2024din}, MAC-SQL~\cite{wang2025mac}, DAIL-SQL~\cite{gao2023text}, and MCS-SQL~\cite{lee2024mcs}. For the open-source LLM baselines, we use the six LLMs mentioned earlier in a zero-shot setting. The fine-tuning-based baselines are represented by specialized LLMs, including CodeS \cite{li2024codes}, ROUTE \cite{qin2025route}, DTS-SQL \cite{dtssq12024}, SENSE \cite{yang2024synthesizing}, OmniSQL \cite{li2025omnisql}, as well as the six base LLMs fine-tuned on five training sets using the Llama-Factory \cite{zheng2024llamafactory} framework. Finally, for the self-play-based methods, we compare with SPIN \cite{chen2024selfplay}. For fairness, we reproduce several baselines using open-source repositories and conduct rigorous evaluations.


\subsection{Comparison Results}

Table \ref{tab:combined_performance} presents the performance of our method and baselines across various datasets, including the SPIDER development and test sets, BIRD development set, SPIDER-SYN, SPIDER-Realistic, and SPIDER-DK. Due to time constraints, we were unable to provide results for our SPFT-SQL on the BIRD test set. In nearly all cases, our SPFT-SQL achieves the best performance. From the results, it is evident that fine-tuning-based methods significantly improve the performance of open-source LLMs. Notably, specialized LLMs (e.g., ROUTE, SENSE, CodeS) fine-tuned on synthetic data outperform those fine-tuned on the original training set but still do not match the performance of closed-source LLMs (e.g., GPT-4). This indicates that the quality of synthetic data remains a limiting factor. The accuracy of the existing self-play method, SPIN, is not only lower than fine-tuning-based methods but also below that of the original model. This is due to SPIN's failure to generate new data for the text-to-SQL task, leading to overfitting from repeated fine-tuning. In contrast, our SPFT-SQL iteratively synthesizes new data during fine-tuning and enhances data quality through a verification feedback mechanism, thereby improving model performance. As a result, our method not only improves the performance of self-play on text-to-SQL tasks but also surpasses specialized LLMs fine-tuned on synthetic data. Particularly on the SPIDER test set, our SPFT-SQL outperforms several existing prompting-based methods (e.g., MAC-SQL), achieving an EX score of 89.1\%, significantly narrowing the gap with GPT-4-based methods.We also present results of our method under different hardness (Appendix ~\ref{Appendix:hardness}), with larger model sizes (Appendix ~\ref{Appendix:larger_llm}), and across more comparison settings (Appendix ~\ref{Appendix:identical_models} and ~\ref{Appendix:more_methods}).

\begin{figure}[t]
  \includegraphics[width=0.48\textwidth]{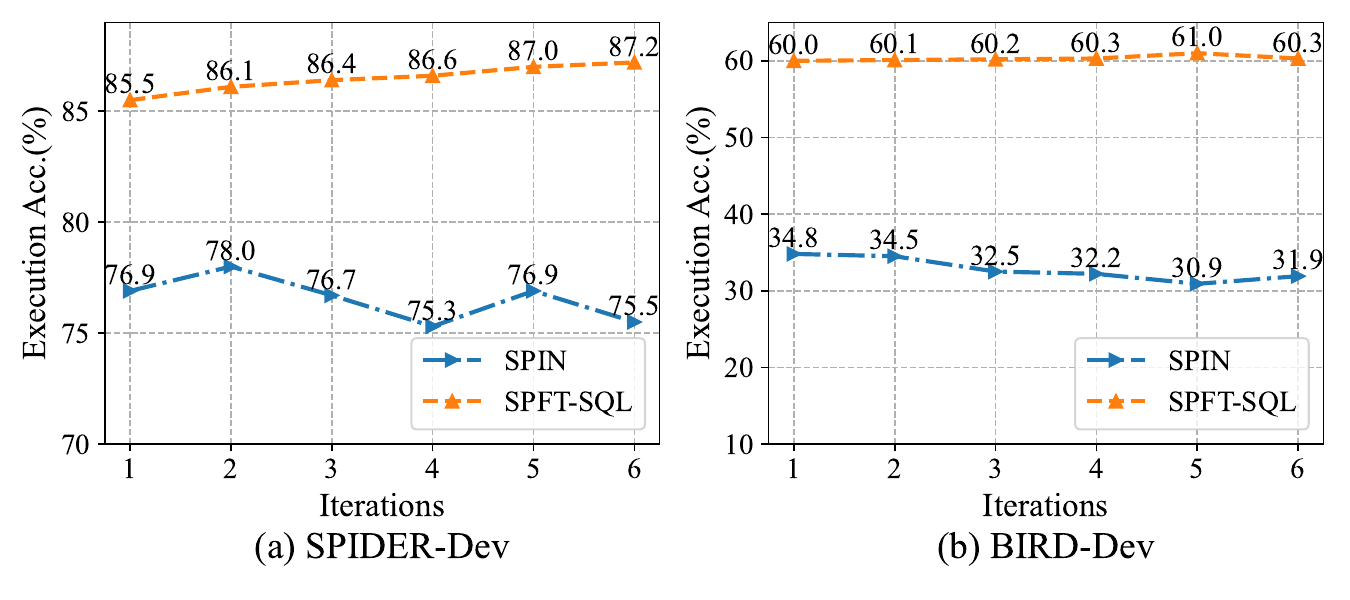}
  \caption {Comparison of Different Iteration}
  \label{fig:iteration}
\end{figure}




\subsection{Parameter Study}

Figure \ref{fig:iteration} shows the performance of SPIN and SPFT-SQL across different iteration rounds on the SPIDER and BIRD development sets, using the Qwen2.5 Coder-7B model as the base model. Our findings reveal that as the number of training iterations increases, the accuracy of SPIN continuously decreases, indicating that SPIN suffers from overfitting with more iterations. In contrast, our method shows a steady improvement in accuracy as iterations progress, suggesting that it continues to generate high-quality data, enhancing model performance and gradually converging. Detailed experimental results can be found in Appendix \ref{Appendix:Quantity}.


\begin{figure}[t]
  \includegraphics[width=0.48\textwidth]{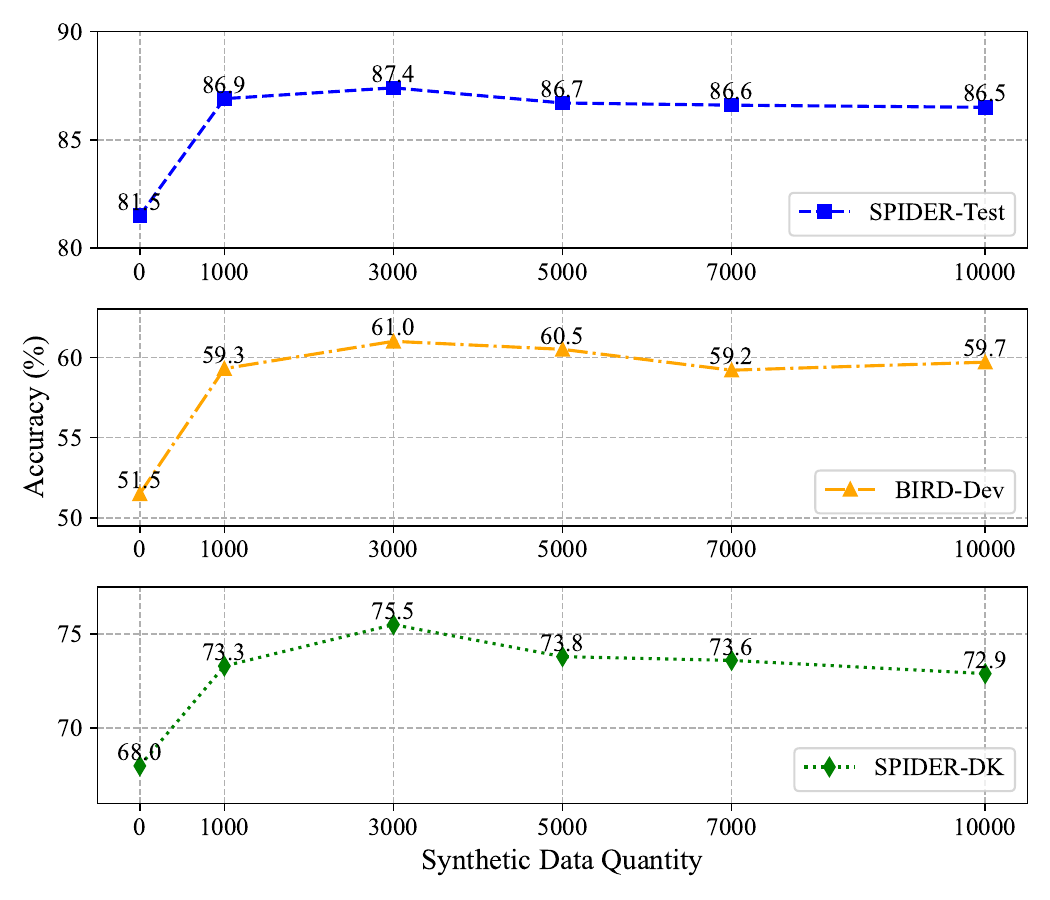}
  \caption {Performance on varing number of synthetic data each round.}
  \label{fig:data_quantity}
\end{figure}

\subsection{Synthetic Data Study}
\paragraph{Synthetic Data Quantity}

As shown in Figure \ref{fig:data_quantity}, we evaluated the impact of varying amounts of synthetic data on model performance at each round during the Verification-Based Iterative Fine-Tuning phase, using the Qwen2.5 Coder-7B model. The results indicate that generating 3,000 data records yields the best performance. This suggests that selecting the appropriate amount of synthetic data during training involves a trade-off: too little data may result in undertraining, while generating excessive data could incur unnecessary time costs.


\paragraph{Synthetic Data Quality}

\begin{figure}[t]
  \includegraphics[width=0.49\textwidth]{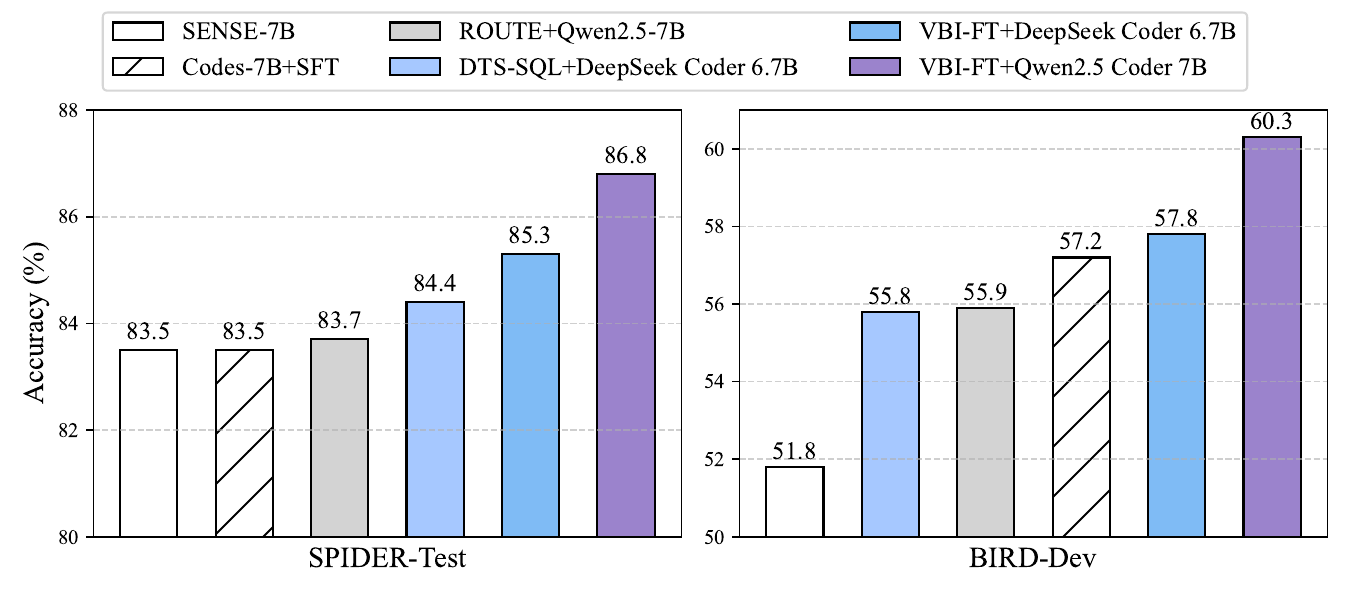}
  \caption {Performance of LLMs fine-tuned on synthetic data generated by SPFT-SQL and baselines.}
  \label{fig:Comparison of Synthetic Data Quality}
\end{figure}

Figure~\ref{fig:Comparison of Synthetic Data Quality} compares LLMs fine-tuned with our Verification-Based Iterative Fine-Tuning (VBI-FT) synthetic data against CodeS, ROUTE, SENSE, and DTS-SQL on SPIDER test and BIRD dev sets.The results demonstrate that SPFT-SQL significantly improves model performance by synthesizing higher-quality data. Using the same base model, SPFT-SQL boosts performance by 3.1\% and 0.9\% on the SPIDER test set compared to ROUTE and DTS-SQL, and by 4.4\% and 2\% on the BIRD development set. Compared to SENSE and CodeS, SPFT-SQL shows notable improvements ranging from 3.3\% to 8.4\%. This improvement can be attributed to the fact that methods like ROUTE and DTS-SQL rely on basic open-source models for data synthesis, which limits the quality of generated data due to the models' inherent capabilities. In contrast, SPFT-SQL overcomes these limitations by leveraging an iterative evaluation feedback mechanism, enhancing the quality of synthetic data and the performance of the model.Appendix ~\ref{Appendix:question_generation} presents a semantic consistency analysis of the synthetic data.


\begin{table}[t]
\centering
\small 
\setlength{\abovecaptionskip}{2pt} 
\setlength{\belowcaptionskip}{2pt}
\setlength{\tabcolsep}{2pt} 
\begin{tabular}{lccccc}
\toprule
 & \multicolumn{3}{c}{\textbf{SPIDER}} & \multicolumn{2}{c}{\textbf{Bird-Dev}} \\
\cmidrule(lr){2-4} \cmidrule(lr){5-6}
 & \textbf{Dev-EX} & \textbf{Dev-TS} & \textbf{Test-EX} & \textbf{EX} & \textbf{VES} \\
\midrule
Qwen2.5 Coder-7B & 83.5 & 79.2 & 81.5 & 51.5 & 55.3 \\
SPFT-SQL & 87.2 & 81.3 & 87.4 & 61.0 & 67.0 \\
w/o VBI-FT & 83.9 & 79.3 & 83.6 & 54.2 & 57.5 \\
w/o Self-Play & 86.6 & 80.4 & 86.8 & 60.3 & 65.3 \\
\bottomrule
\end{tabular}
\caption{The ablation study results on the SPIDER and the BIRD datasets.}
\label{tab:ablation_study}
\end{table}

\subsection{Ablation Study}

As shown in Table \ref{tab:ablation_study}, we conducted an ablation study on the SPIDER and BIRD datasets using Qwen2.5 Coder-7B as the base LLM. The results reveal two key findings. First, Verification-Based Iterative Fine-Tuning (VBI-FT) improved performance by 3.3\% to 3.8\% in EX on the SPIDER dataset and by 6.8\% on the BIRD dataset, highlighting its significance in enhancing core SQL synthesis abilities. Additionally, the self-play fine-tuning process resulted in an accuracy boost of 0.6\% to 0.8\% on both datasets, demonstrating that self-play enables the model to leverage its intrinsic capabilities without external supervision. Together, these findings underscore the effectiveness of both Verification-Based Iterative Fine-Tuning and self-play fine-tuning in improving model performance on text-to-SQL tasks. Additional results for the SPIDER-variant can be found in Appendix \ref{Appendix:Ablation}.


\section{Conclusion}

In this paper, we propose a novel self-play fine-tuning framework for text-to-SQL tasks, called SPFT-SQL. SPFT-SQL enhances the capabilities of open-source models through verification-based iterative fine-tuning to generate high-quality data augmentation, while further improving the models' ability to generate accurate SQL via error-driven adversarial training in self-play scenarios. Our work represents the first effective implementation of the self-play method in text-to-SQL tasks, significantly narrowing the performance gap between open-source and closed-source models. Future research will focus on exploring cross-domain generalization capabilities and developing efficient adversarial architectures.


\section*{Limitations}

Although our method has shown promising performance and significant progress across various aspects, there are several limitations and areas for further improvement. Firstly, the introduction of data synthesis and additional evaluation steps inevitably introduces some extra computational overhead during training. Secondly, while our current data generation and filtering strategies have proven effective, there is still room for further exploration in data selection techniques, which may lead to improvements in the overall quality and relevance of the generated data.

\section*{Acknowledgments}
This study is supported by the National Key Research and Development Program of China under Grant 2023YFB3106504, Guangdong Provincial Key Laboratory of Novel Security Intelligence Technologies under Grant 2022B1212010005, the China Postdoctoral Science Foundation under Grant Number 2024M751555, the Major Key Project of PCL under Grant PCL2024A04, Shenzhen Science and Technology Program under Grant ZDSYS20210623091809029 and RCBS20221008093131089.


\bibliography{custom}

\appendix

\section{Appendix}
\label{sec:appendix}
\subsection{Prompt}
\begin{figure*}[ht]
  \includegraphics[width=1\textwidth]{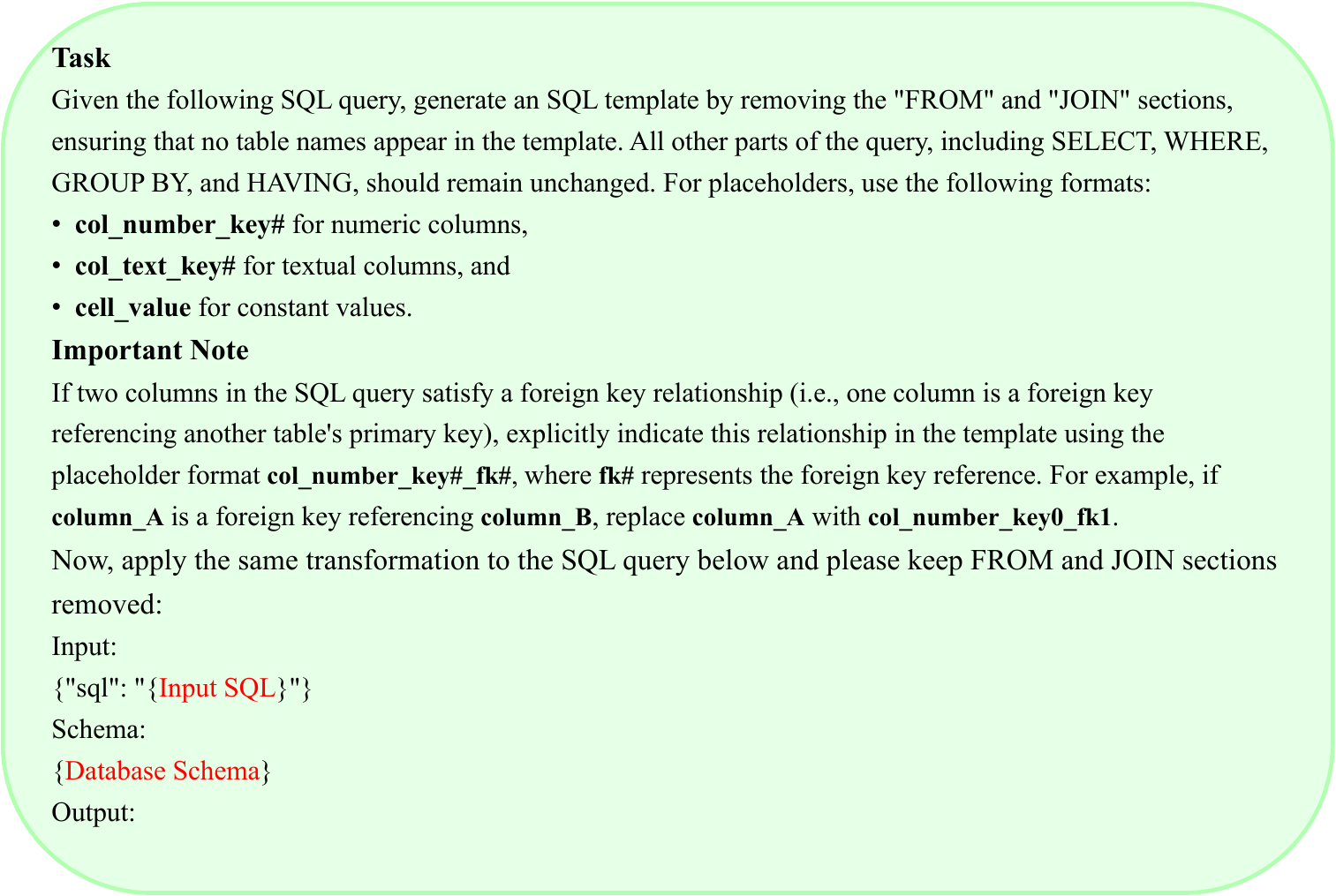}
  \caption {Prompt for extracting standardized SQL templates.}
  \label{fig:Template_Extraction}
\end{figure*}

In this section, we provide the prompts employed for the methodology described under Template Extraction, which are depicted in Figure \ref{fig:Template_Extraction}.
\label{sec:Template_Extraction_prompt}

\subsection{Theoretical Analysis and Comparison with DPO}
\label{Appendix:Dpo_Analysis}

To clarify the theoretical distinctions and advantages of our SPFT-SQL method relative to Direct Preference Optimization (DPO)~\cite{rafailov2024direct}, we provide a detailed mathematical explanation below.

\paragraph{Error-driven Reward Signal}
The standard DPO loss function is defined as follows:
\begin{align}
\mathcal{L}_{\text{DPO}}(\pi_\theta;\pi_{\mathrm{ref}}) = 
-\mathbb{E}&\left[\log\sigma\left(\beta\log\frac{\pi_\theta(y_w|x)}{\pi_{\mathrm{ref}}(y_w|x)} \right.\right. \notag \\ 
&\quad \left.\left. - \beta\log\frac{\pi_\theta(y_l|x)}{\pi_{\mathrm{ref}}(y_l|x)}\right)\right],
\end{align}
where $y_w$ denotes the preferred query and $y_l$ denotes the less preferred query. This formulation uses relative log-probabilities and assumes a clear preference signal based solely on output comparison, without incorporating task-specific execution feedback.However, the Text-to-SQL task presents unique challenges:

\textit{1.Semantic Equivalence}: SQL queries differing significantly in syntax can produce identical execution results, resulting in negligible probability differences and weakening gradient signals.

\textit{2.Local Optima}: When model outputs closely approximate those of a reference model, the gradient signal diminishes, trapping optimization in local optima.

To address these limitations, we introduce an error-driven reward signal based explicitly on SQL execution results in Equation (\ref{eq:reward_signal}),  and our self-play loss function is defined as Equation (\ref{eq:self_play_loss}).This modified loss provides two primary theoretical enhancements:
\paragraph{1. Amplified Error Signals}
By incorporating execution-based rewards, we explicitly amplify gradient updates for incorrect SQL predictions, providing a strong corrective signal, even when queries are semantically similar.
\paragraph{2. Improved Gradient Stability}
For DPO, when preferred and non-preferred outputs become nearly indistinguishable, the gradient difference ($\Delta r$) tends toward zero:
\[
\Delta r = \beta\log\frac{\pi_\theta(y_w|x)}{\pi_{\mathrm{ref}}(y_w|x)}-\beta\log\frac{\pi_\theta(y_l|x)}{\pi_{\mathrm{ref}}(y_l|x)} \approx 0,
\]
resulting in weak updates. Our error-driven approach prevents gradient vanishing by introducing the binary execution-based reward $R(y', x)$, resulting in the following gradient update:

\begin{align}&
\nabla_{\theta} \mathcal{L}_{\text{Self-Play}} 
\propto -\lambda R(y', x) \, \sigma(\Delta r) \bigg[ \notag\\
&\quad \nabla_{\theta} \log p_{\theta_m}(y_+|x)
     - \nabla_{\theta} \log p_{\theta_m}(y_-|x) \bigg]
\end{align}
Thus, even small probability differences between queries yield meaningful updates if they differ in execution correctness.

\paragraph{Dynamic Reference Model Update}
Standard DPO employs a static reference model, causing stagnation as the optimized model nears the reference. To avoid local optima, we dynamically update our reference (opponent) model $p_{\theta_{o}}$ iteratively:
\[
p_{\theta_{o}}^{(t+1)} \leftarrow \arg\min_{p_{\theta}\in M'}Acc(p_{\theta}, D_{val})
\]
where $M'$ excludes the previously used opponent. This ensures continual competitive pressure, progressively guiding the policy model away from local optima and toward improved global solutions.

From a theoretical perspective, our dynamic approach aligns closely with DPO principles. Specifically, under KL-constraints, the optimal policy for DPO is defined as:
\[
\pi^*(y|x) = \frac{1}{Z(x)}\pi_{\mathrm{ref}}(y|x)\exp\left(\frac{1}{\beta}r(x,y)\right)
\]

In our dynamic setting, iterative refinement leads toward a similarly defined global optimum:
\[
\pi^*(y|x) = \frac{1}{Z(x)}p_{\theta_o}(y|x)\exp\left(\frac{1}{\beta}R(y,x)\right)
\]
achieved through successive improvements in reference models and execution-based feedback.

In summary, our SPFT-SQL method provides the following theoretical and practical advantages over DPO:
\begin{itemize}
    \setlength\itemindent{0pt}
    \setlength\leftskip{0pt}
    \item \textbf{Explicit Execution-based Rewards}: Directly penalize SQL execution errors, ensuring robust gradient signals.
    \item \textbf{Dynamic Opponent Update}: Iteratively improve the reference model, preventing optimization stagnation.
    \item \textbf{Theoretical Alignment}: Maintain consistency with DPO theory while addressing practical challenges inherent to Text-to-SQL tasks.
\end{itemize}

These theoretical considerations, supported by empirical results (Table~\ref{tab:dpo_ablation}), clearly demonstrate the advantages of our proposed methodology compared to conventional DPO-based approaches.

\subsection{Ablation Study with DPO and PPO}
\label{Appendix:dpo_ppo_ablation}

To further analyze the effectiveness of our SPFT-SQL method, we conducted an ablation study comparing our method with alternative fine-tuning strategies, including Direct Preference Optimization (DPO)~\cite{rafailov2024direct} and Proximal Policy Optimization (PPO)~\cite{schulman2017proximal}. Although our approach does not directly utilize DPO, this comparison clarifies the advantages of our design, particularly our error-driven feedback mechanism.

We evaluated various optimization strategies on the SPIDER dataset using the Qwen2.5 Coder-7B model. The performance comparison, summarized in Table~\ref{tab:dpo_ablation}, includes the baseline model, standard implementations of DPO and PPO, as well as combined approaches incorporating VBI-FT (Verification-Based Iterative Fine-Tuning). 

\begin{table}[ht]
\centering
\small 
\setlength{\abovecaptionskip}{2pt} 
\setlength{\belowcaptionskip}{2pt}
\setlength{\tabcolsep}{2pt} 
\begin{tabular}{lcc}
\toprule
\textbf{Method}                        & \textbf{SPIDER Dev} & \textbf{SPIDER Test} \\ 
\midrule
Qwen2.5 Coder-7B               & 83.5                     & 81.5                      \\
+ DPO                 & 79.4                     & 78.2                      \\
+ PPO                 & 81.8                     & 83.7                      \\
+ VBI-FT + DPO        & 84.7                     & 84.6                      \\
+ VBI-FT + PPO        & 85.1                     & 86.6                      \\
+ \textbf{SPFT-SQL} & \textbf{87.2}            & \textbf{87.4}             \\ 
\bottomrule
\end{tabular}
\caption{Ablation Study Comparing SPFT-SQL with DPO and PPO Methods on SPIDER dataset (EX)}
\label{tab:dpo_ablation}
\end{table}

Experimental results clearly indicate the superiority of SPFT-SQL over both DPO- and PPO-based approaches. The primary advantage of SPFT-SQL originates from its explicit error-driven loss function, which penalizes SQL samples leading to incorrect predictions by main model and reinforces correctly executed SQL queries. In contrast:

\begin{itemize}
    \item \textbf{DPO}, lacking direct SQL execution feedback, mistakenly penalizes correctly predicted SQL queries when treated as negative samples, thus negatively affecting overall performance.
    \item \textbf{PPO} depends heavily on reward or value models to evaluate generated SQL statements during policy updates. However, accurately assessing the quality of SQL queries solely through natural language evaluation (without direct execution) remains significantly challenging.
\end{itemize}

This ablation study highlights the crucial role played by our error-driven loss function, enabling SPFT-SQL to effectively leverage direct execution-based evaluations, leading to enhanced performance on text-to-SQL tasks.

\subsection{Experimental Design and Dataset Details}

\label{Appendix:exp_design}

In order to comprehensively evaluate the robustness and generalization capabilities of the proposed SPFT-SQL method, experiments are conducted across five distinct and challenging cross-domain datasets. These datasets are commonly employed benchmarks within Text-to-SQL research: SPIDER~\cite{yu2018spider}, BIRD~\cite{li2024can}, SPIDER-Syn~\cite{gan2021towards}, SPIDER-Realistic~\cite{deng2021realistic}, and SPIDER-DK~\cite{gan2021dk}. The complexity and diversity of these datasets facilitate a rigorous assessment of model generalization performance.

\paragraph{Dataset Statistics}
The SPIDER dataset is recognized for its diversity, encompassing 206 databases spanning 138 distinct domains, with a total of 876 tables and 4,669 columns. BIRD, on the other hand, is noted for its emphasis on domain-specific complexity, consisting of 597 tables and 4,417 columns across 80 databases from 37 unique domains. Additional SPIDER-based datasets (SPIDER-Syn, SPIDER-Realistic, and SPIDER-DK) are created through transformations aimed at evaluating robustness under variations such as synonym substitution, more realistic question formulations, and knowledge-augmented queries. Table~\ref{tab:dataset_stats} provides detailed statistics for each benchmark.

\begin{table*}[t]
\centering
\small 
\begin{tabular}{lccccc}
\toprule
Dataset          & Databases           & Domains & Tables & Columns & Metric           \\ 
\midrule
Spider           & 166 train + 40 test & 138     & 876    & 4,669   & EX               \\
Spider-Syn       & 20                  & 20      & 85     & 452     & EX+TS            \\
Spider-Realistic & 19                  & 19      & 81     & 435     & EX+TS            \\
Spider-DK        & 6                   & 6       & 49     & 272     & EX               \\
Bird             & 69 train + 11 dev   & 37      & 597    & 4,417   & EX+VES           \\ 
\bottomrule
\end{tabular}
\caption{Summary of Benchmark Dataset Statistics}
\label{tab:dataset_stats}
\end{table*}

\paragraph{Fine-Tuning Setup}
For fine-tuning-based approaches utilizing open-source large language models (LLMs), models such as CodeS~\cite{li2024codes}, ROUTE~\cite{qin2025route}, DTS-SQL~\cite{dtssq12024}, and SENSE~\cite{yang2024synthesizing} followed their originally proposed methodologies. The additional models, namely Llama3-8B, DeepSeek-7B, and various sizes of Qwen2.5 Coder (1.5B, 7B, 14B, and 32B), were fine-tuned using supervised fine-tuning (SFT) protocols with Low-Rank Adaptation~\cite{hu2022lora} through the Llama-Factory framework~\cite{zheng2024llamafactory}. The specific hyperparameters included the AdamW optimizer, a batch size of 64, a learning rate of $2\times10^{-4}$, and fine-tuning for three epochs on each training set. This consistent LoRA-based approach ensures computational efficiency as well as fair and reproducible comparisons across models and datasets.

Overall, these comprehensive experiments and clearly defined fine-tuning strategies reinforce the evaluation rigor and robustness of our SPFT-SQL methodology, demonstrating its superior generalization performance across diverse and challenging benchmarks.

\subsection{Comparison with different hardness}
\label{Appendix:hardness}

To comprehensively evaluate the model performance, we adopted methodologies from pertinent studies \cite{pourreza2024din,gao2023text,qin2025route} and computed the EX score on the development sets of SPIDER and BIRD. The results presented in Table \ref{tab:hardness_performance_spider} and Table \ref{tab:hardness_performance_BIRD} demonstrate that the SPFT-SQL approach excels both in overall performance and across various difficulty levels, thereby further validating the efficacy of our proposed method.

\begin{table*}[h!]
\centering
\small
\begin{tabular}{lccccc}
\toprule
\textbf{Method} & \textbf{Easy} & \textbf{Medium} & \textbf{Hard} & \textbf{Extra} & \textbf{All} \\
\midrule
\textit{Prompting with GPT} & & & & & \\
DIN-SQL+GPT4\cite{pourreza2024din} & 92.3 & 87.4 & 76.4 & 62.7 & 82.8 \\
DAIL-SQL+GPT4\cite{gao2023text} & 91.5 & 90.1 & 75.3 & 62.7 & 83.6 \\
MCS-SQL+GPT4\cite{lee2024mcs} & 94.0 & 93.5 & 88.5 & 72.9 & 89.5 \\
\midrule
\textit{Fine-Tuning with Open-Source LLMs} & & & & & \\
Codes-7B+SFT\cite{li2024codes}  & 94.8 & 91.0 & 75.3 & 66.9 & 85.4 \\
Codes-15B+SFT\cite{li2024codes} & 95.6 & 90.4 & 78.2 & 61.4 & 84.9 \\
SENSE-7B\cite{yang2024synthesizing} & 95.2 & 88.6 & 75.9 & 60.3 & 83.5 \\
ROUTE+Qwen2.5-7B\cite{qin2025route} & 92.8 & 89.7 & 77.0 & 60.2 & 83.6 \\
ROUTE+Qwen2.5-14B\cite{qin2025route} & 94.0 & 93.0 & 81.6 & \textbf{68.1} & 87.3 \\
\midrule
\textit{Self-Play Method} & & & & & \\
SPIN+Qwen2.5 Coder-14B\cite{chen2024selfplay} & 91.5 & 87.7 & 74.7 & 63.3 & 82.3 \\
\textbf{SPFT-SQL+Qwen2.5 Coder-1.5B} & 92.3 & 83.4 & 72.4 & 58.4 & 79.7 \\
\textbf{SPFT-SQL+Qwen2.5 Coder-7B} & 96.4 & 91.9 & \textbf{85.1} & 62.7 & 87.2 \\
\textbf{SPFT-SQL+Qwen2.5 Coder-14B} & 95.6 & \textbf{94.4} & 81.6 & 64.5 & 87.7 \\
\textbf{SPFT-SQL+Qwen2.5 Coder-32B} & \textbf{96.4} & 93.5 & 80.5 & 67.5 & \textbf{87.8} \\
\bottomrule
\end{tabular}
\caption{The performance (EX) comparison with different hardness on the SPIDER-Dev}
\label{tab:hardness_performance_spider}
\end{table*}

\begin{table*}[h!]
\centering
\small
\begin{tabular}{lcccc}
\toprule
\textbf{Method} & \textbf{Simple} & \textbf{Moderate} & \textbf{Challenging} & \textbf{All} \\
\midrule
\textit{Prompting with GPT} & & & & \\
MAC-SQL+GPT4\cite{wang2025mac} & 65.7 & 52.7 & 40.3 & 59.4 \\
MCS-SQL+GPT4\cite{lee2024mcs} & 70.4 & 53.1 & 51.4 & 63.4 \\
\midrule
\textit{Fine-Tuning with Open-Source LLMs} & & & & \\
Codes-7B+SFT\cite{li2024codes} & 64.6 & 46.9 & 40.3 & 57.2 \\
Codes-15B+SFT\cite{li2024codes} & 65.8 & 48.8 & 42.4 & 58.5 \\
ROUTE+Qwen2.5-7B\cite{qin2025route} & 63.8 & 45.4 & 39.6 & 55.9 \\
ROUTE+Qwen2.5-14B\cite{qin2025route} & 67.7 & 53.1 & 42.4 & 60.9 \\
\midrule
\textit{Self-Play Method} & & & & \\
SPIN+Qwen2.5 Coder-14B\cite{chen2024selfplay} & 45.8 & 24.1 & 20.1 & 36.8 \\
\textbf{SPFT-SQL+Qwen2.5 Coder-1.5B} & 61.1 & 46.5 & 33.3 & 54.0 \\
\textbf{SPFT-SQL+Qwen2.5 Coder-7B} & 68.7 & 51.6 & 41.7 & 61.0 \\
\textbf{SPFT-SQL+Qwen2.5 Coder-14B} & 68.8 & \textbf{57.6} & 49.3 & 63.6 \\
\textbf{SPFT-SQL+Qwen2.5 Coder-32B} & \textbf{71.2} & 57.4 & \textbf{51.4} & \textbf{65.2} \\
\bottomrule
\end{tabular}
\caption{The performance (EX) comparison with different hardness on the BIRD-Dev}
\label{tab:hardness_performance_BIRD}
\end{table*}

\begin{table*}[h!]
\centering
\small
\begin{tabular}{lccccccccccc}
\toprule
& \multicolumn{3}{c}{\textbf{SPIDER}} & \multicolumn{5}{c}{\textbf{SPIDER-Variants}} & \multicolumn{2}{c}{\textbf{BIRD}}\\
 \cmidrule(lr){2-4} \cmidrule(lr){5-9} \cmidrule(lr){10-11}
\textbf{Quantity} & \multicolumn{2}{c}{\textbf{Dev}} & \textbf{Test} & \multicolumn{2}{c}{\textbf{Syn}} & \multicolumn{2}{c}{\textbf{Realistic}} & \textbf{DK} & \multicolumn{2}{c}{\textbf{Dev}} \\
\cmidrule(lr){2-3} \cmidrule(lr){5-6} \cmidrule(lr){7-8} \cmidrule(lr){10-11}
 & \textbf{EX} & \textbf{TS} & \textbf{EX} & \textbf{EX} & \textbf{TS} & \textbf{EX} & \textbf{TS} & \textbf{EX} & \textbf{EX} & \textbf{VES} \\
\midrule
 0    & 83.5    & 79.2    & 81.5    & 69.8    & 64.2    & 75.4    & 70.9    & 68.0  & 51.5    & 55.3  \\
1000   & 86.5    & 80.9    & 86.9     & \textbf{76.8}    & \textbf{69.6}    & 82.1    & \textbf{76.6}    & 73.3  & 59.3    & 62.7  \\
3000   & \textbf{87.2}    & \textbf{81.3}    & \textbf{87.4}    & 75.1    & 67.6    & \textbf{83.3}    & 75.6    & \textbf{75.5}   & \textbf{61.0}    & \textbf{67.0} \\
5000   & 86.3    & 79.6    & 86.7    & 72.1    & 64.3    & 82.8    & 75.6    & 73.8    & 60.5    & 66.2 \\
7000   & 86.6    & 80.9    & 86.6    & 71.8    & 64.1    & 82.1    & 73.6    & 73.6   & 59.2    & 65.8 \\
10000  & 86.0    & 80.4    & 86.5    & 72.4    & 64.6    & 82.5    & 73.6    & 72.9   & 59.7    & 62.9 \\
\bottomrule
\end{tabular}
\caption{Comparison of Synthetic Data Quantity}
\label{tab:Quantity_full}
\end{table*}

\subsection{Comparison of Synthetic Data Quantity}
\label{Appendix:Quantity}
Table \ref{tab:Quantity_full} presents the experimental results across all datasets for different amounts of synthetic data.The experimental results show that the model achieves the best performance when generating 3,000 synthetic data records. Specifically, it achieves an accuracy of 87.4\% on the SPIDER-Test dataset, 87.2\% on the SPIDER-Dev dataset, 61.0\% on the BIRD-Dev dataset. These results indicate that generating an appropriate amount of synthetic data is crucial for improving model performance.

When the amount of generated data is relatively small, the model's performance improves but does not reach its optimal state. For example, the accuracy on the SPIDER-Realistic dataset is 82.1\%, on the SPIDER-DK dataset is 73.3\%, and on the BIRD-Dev dataset is 59.3\%. This suggests that insufficient data may prevent the model from learning enough information, thereby limiting its performance.

On the other hand, when the amount of generated data is excessive, the model's performance declines. For instance, when generating 5,000 records, the accuracy on the SPIDER-Test dataset drops to 86.7\%, on the SPIDER-Syn dataset to 72.1\%, and on the BIRD-Dev dataset to 60.5\%. This could incur unnecessary time costs.

In conclusion, generating 3,000 synthetic data records is an ideal choice, as it ensures data quality while maximizing model performance improvement. This finding emphasizes the importance of selecting an appropriate amount of synthetic data during training to avoid compromising the model's final performance due to insufficient or excessive data.

\subsection{Ablation Study Results}
\label{Appendix:Ablation}
Table \ref{tab:Ablation_full} delineates the ablation study results across all datasets, shedding light on the individual contributions of various components to the overall system performance. The implementation of Verification-Based Iterative Fine-Tuning (VBI-FT) significantly enhanced the model's performance, with improvements ranging from 5.2\% to 8.3\% in EX on the SPIDER-Variants, highlighting its critical role in advancing core SQL synthesis capabilities. Furthermore, the self-play fine-tuning process contributed to an accuracy increase of 0.3\% to 1.7\% on the SPIDER-Variants, illustrating how self-play allows the model to optimize its inherent potential without relying on external supervision. Collectively, these results underscore the efficacy of both Verification-Based Iterative Fine-Tuning and self-play fine-tuning in boosting the model's performance on SQL synthesis tasks.

\begin{table*}[h!]
\centering
\small
\begin{tabular}{lccccccccccc}
\toprule
& \multicolumn{3}{c}{\textbf{SPIDER}} & \multicolumn{5}{c}{\textbf{SPIDER-Variants}} & \multicolumn{2}{c}{\textbf{BIRD}}\\
 \cmidrule(lr){2-4} \cmidrule(lr){5-9} \cmidrule(lr){10-11}
 & \multicolumn{2}{c}{\textbf{Dev}} & \textbf{Test} & \multicolumn{2}{c}{\textbf{Syn}} & \multicolumn{2}{c}{\textbf{Realistic}} & \textbf{DK} & \multicolumn{2}{c}{\textbf{Dev}} \\
\cmidrule(lr){2-3} \cmidrule(lr){5-6} \cmidrule(lr){7-8} \cmidrule(lr){10-11}
 & \textbf{EX} & \textbf{TS} & \textbf{EX} & \textbf{EX} & \textbf{TS} & \textbf{EX} & \textbf{TS} & \textbf{EX} & \textbf{EX} & \textbf{VES} \\
\midrule
 Qwen2.5 Coder-7B    & 83.5    & 79.2    & 81.5    & 69.8    & 64.2    & 75.4    & 70.9    & 68.0  & 51.5    & 55.3  \\
 SPFT-SQL   & 87.2    & 81.3    & 87.4     & 75.1    & 67.6    & 83.3    & 75.6    & 75.5  & 61.0    & 67.0  \\
 w/o VBI-FT   & 83.9    & 79.3    & 83.6    & 69.9    & 64.6    & 75.0    & 70.5    & 70.3   & 54.2    & 57.5 \\
 w/o self-play & 86.6    & 80.4    & 86.8    & 74.8    & 66.8    & 82.7    & 74.4    & 73.8    & 60.3    & 65.3 \\
\bottomrule
\end{tabular}
\caption{Ablation Study Results}
\label{tab:Ablation_full}
\end{table*}

\subsection{Scalability Analysis for Larger LLMs}
\label{Appendix:larger_llm}

To assess the scalability and robustness of our SPFT-SQL method beyond previously evaluated model sizes, we conducted further experiments employing a significantly larger model, \textbf{Llama3-70B-Instruct}, on the widely-used SPIDER dataset~\cite{yu2018spider}. Previously evaluated models included parameter scales of 1.5B, 7B, 14B, and 32B. Extending our analysis to the 70B scale allows us to better understand the performance trends and potential convergence at increased parameter scales.

\begin{table}[ht]
\centering
\small 
\setlength{\abovecaptionskip}{2pt} 
\setlength{\belowcaptionskip}{2pt}
\setlength{\tabcolsep}{8pt} 
\begin{tabular}{lcc}
\toprule
\textbf{Method}                     & \textbf{Dev} & \textbf{Test} \\ 
\midrule
SPFT-SQL + Qwen2.5 Coder 32B        & 87.8                     & 89.1                      \\
SFT + Llama3-70B-Instruct           & 82.1                     & 84.6                      \\
SPIN + Llama3-70B-Instruct          & 81.6                     & 83.2                      \\
SPFT-SQL + Llama3-70B-Instruct      & 85.4                     & 88.7                      \\
\bottomrule
\end{tabular}
\caption{Scalability Analysis of SPFT-SQL on SPIDER dataset (EX)}
\label{tab:scalability_results}
\end{table}

Table~\ref{tab:scalability_results} summarizes the comparative performance of SPFT-SQL using the 70B-scale model alongside previously tested approaches such as SFT and SPIN. These experiments demonstrate that SPFT-SQL continues to outperform the baseline methods at larger model scales, maintaining its relative advantage. Notably, the results indicate convergence tendencies in Text-to-SQL task performance, as evidenced by the comparable outcomes of the Qwen2.5 Coder 32B and Llama3-70B-Instruct models.These results confirm the scalability of our approach and suggest that improvements obtained by scaling to substantially larger models, beyond 32B parameters, may exhibit diminishing returns. Nevertheless, our SPFT-SQL maintains clear performance superiority, affirming its robustness and effectiveness at scale.

\subsection{Comparative Evaluation with Identical Base Models}
\label{Appendix:identical_models}

To further ensure a fair and rigorous evaluation of our proposed SPFT-SQL method, we conducted additional experiments comparing SPFT-SQL directly against other representative methods—namely, SENSE~\cite{yang2024synthesizing} and ROUTE~\cite{qin2025route}—using identical base models. The results clarify the contributions of our method beyond merely leveraging stronger open-source base models.

\paragraph{Comparison with SENSE on CodeLlama-7B}

We first evaluated our approach against SENSE using the same foundational model, CodeLlama-7B, across two widely used benchmarks, SPIDER and BIRD. Unlike SENSE, which leverages closed-source models for data generation and a single round of Direct Preference Optimization (DPO), SPFT-SQL employs iterative VBI-FT and a dynamic, error-driven self-play framework. The comparative results are shown in Table~\ref{tab:sense_comparison}.

\begin{table}[ht]
\centering
\small 
\setlength{\abovecaptionskip}{2pt} 
\setlength{\belowcaptionskip}{2pt}
\setlength{\tabcolsep}{2pt} 
\begin{tabular}{lccc}
\toprule
\textbf{Method} & \textbf{SPIDER Dev} & \textbf{SPIDER Test} & \textbf{BIRD Dev} \\ 
\midrule
CodeLlama-7B           & 61.1                     & 48.3                      & 17.9                   \\
+ SFT             & 71.5                     & 72.3                      & 40.2                   \\
+ SENSE           & 83.2                     & 83.5                      & 51.8                   \\
+ \textbf{SPFT-SQL}       & \textbf{83.4}            & \textbf{84.1}             & \textbf{53.4}          \\ 
\bottomrule
\end{tabular}
\caption{Performance Comparison with SENSE using CodeLlama-7B}
\label{tab:sense_comparison}
\end{table}

The results clearly demonstrate that SPFT-SQL consistently outperforms SENSE, even under identical base-model conditions, affirming our framework's superior capability to enhance Text-to-SQL performance without relying on closed-source models.

\paragraph{Comparison with ROUTE on Qwen 2.5-7B}

We further compared SPFT-SQL with ROUTE using Qwen 2.5-7B, addressing concerns regarding base-model discrepancies (e.g., Qwen vs. Qwen-Coder variants). ROUTE specifically leverages the open-source Qwen model without specialized coding pre-training. Therefore, to ensure fairness and transparency, we replicated our SPFT-SQL evaluations using exactly the same Qwen 2.5-7B base model. The results, summarized in Table~\ref{tab:route_comparison}, clearly highlight the relative improvement of our approach.

\begin{table}[ht]
\centering
\small 
\setlength{\abovecaptionskip}{2pt} 
\setlength{\belowcaptionskip}{2pt}
\setlength{\tabcolsep}{2pt} 
\begin{tabular}{lccc}
\toprule
\textbf{Method} & \textbf{SPIDER Dev} & \textbf{SPIDER Test} & \textbf{BIRD Dev} \\ 
\midrule
Qwen 2.5-7B          & 72.5                     & 75.9                      & 41.1                   \\
  + SFT            & 80.9                     & 82.8                      & 51.4                   \\
  + ROUTE           & 83.6                     & 83.7                      & 55.9                   \\
  + \textbf{SPFT-SQL}      & \textbf{84.9}            & \textbf{85.6}             & \textbf{59.6}          \\ 
\bottomrule
\end{tabular}
\caption{Performance Comparison with ROUTE using Qwen 2.5-7B}
\label{tab:route_comparison}
\end{table}

These experiments explicitly confirm that SPFT-SQL’s superior performance does not merely result from using stronger base models, but rather from the intrinsic benefits of our iterative VBI-FT strategy combined with the dynamic error-driven self-play mechanism. Collectively, these fair-model comparisons substantiate the general applicability and effectiveness of our method across various open-source base models, ensuring broader reproducibility and clear insights for future research applications.

\subsection{Semantic Consistency Analysis for Question Generation}
\label{Appendix:question_generation}

To quantify potential bias introduced during the generation of synthetic validation data, we conducted a semantic consistency experiment. Specifically, we evaluated the semantic alignment between SQL-generated natural language questions (NLQs) and original questions from the SPIDER Dev set. The SQL-to-Text model used for this evaluation was Qwen 2.5-Coder-7B, both before and after iterative optimization.

We randomly sampled 200 NLQ-SQL pairs from the SPIDER Dev set and generated NLQs using the Qwen 2.5-Coder-7B model based solely on SQL queries. To assess semantic consistency, evaluations were performed independently by GPT-4 and human annotators. Table~\ref{tab:semantic_consistency} summarizes these results.

\begin{table}[ht]
\centering
\small
\setlength{\abovecaptionskip}{2pt}
\setlength{\belowcaptionskip}{2pt}
\setlength{\tabcolsep}{2pt}
\begin{tabular}{lcc}
\toprule
\textbf{SQL-to-Text Model} & \textbf{GPT-4 Eval} & \textbf{Human Eval} \\
\midrule
Before optimization       & 112/200 (56.0\%)     & 113/200 (56.5\%)     \\
After optimization        & 184/200 (92.0\%)     & 187/200 (93.5\%)     \\
\bottomrule
\end{tabular}
\caption{Semantic Consistency of Generated NLQs}
\label{tab:semantic_consistency}
\end{table}

The significant improvement observed after iterative optimization (from 56.5\% to 93.5\% human-evaluated consistency)  demonstrates our approach's efficacy in minimizing semantic biases introduced during synthetic validation set construction.

\subsection{Comparison with More Methods}
\label{Appendix:more_methods}

To further benchmark the effectiveness of our SPFT-SQL method, we compared its performance against two recently proposed state-of-the-art approaches: XiYan-SQL~\cite{gao2024xiyan} and CHASE-SQL~\cite{pourreza2024chase}. Table~\ref{tab:sota_comparison} presents these comparative results on both the SPIDER and BIRD datasets.

\begin{table}[htbp]
\centering
\small
\setlength{\abovecaptionskip}{2pt}
\setlength{\belowcaptionskip}{2pt}
\setlength{\tabcolsep}{1.5pt}
\begin{tabular}{lcc}
\toprule
\textbf{Method} & \textbf{SPIDER Test} & \textbf{BIRD Dev} \\
\midrule
CHASE-SQL+Gemini 1.5      & 87.6           & 73.0          \\
XiYan-SQL                   & \textbf{89.6}  & \textbf{73.3} \\
SPFT-SQL+Qwen2.5 Coder-32B & 89.1   & 65.2          \\
\bottomrule
\end{tabular}
\caption{Comparison with Recent Methods (EX)}
\label{tab:sota_comparison}
\end{table}

On the SPIDER dataset, SPFT-SQL demonstrates competitive performance, surpassing CHASE-SQL (Gemini 1.5) and approaching the accuracy of XiYan-SQL. This indicates that our method effectively synthesizes high-quality training data, enabling relatively smaller open-source models to achieve performance closely rivaling large-scale closed-source models.

However, on the more complex BIRD dataset, there remains a performance gap of approximately 7.8\% compared to CHASE-SQL and 8.1\% compared to XiYan-SQL. We attribute this difference to the inherent methodological distinctions: both CHASE-SQL and XiYan-SQL employ advanced multi-agent frameworks, leveraging collaborative interactions among multiple agents to enhance SQL generation capabilities. In contrast, SPFT-SQL primarily focuses on single-model fine-tuning through iterative, adaptive learning without multi-agent collaboration.

These insights provide clear directions for future research, suggesting potential enhancements by integrating multi-agent collaboration within our adaptive fine-tuning framework.


\subsection{Computational Resource and Storage Analysis}
\label{Appendix:computational}

We analyzed the computational resource usage and storage overhead of our SPFT-SQL method relative to standard supervised fine-tuning (SFT) and the self-play approach SPIN~\cite{chen2024selfplay}. All experiments used the Qwen2.5 Coder 7B model on the SPIDER dataset. We utilized 8*NVIDIA A800 GPUs, each equipped with 80GB memory, for fair comparisons.

\begin{table*}[htbp]
\centering
\small
\begin{tabular}{lccccc}
\toprule
\textbf{Method} & \textbf{Time per Iter.} & \textbf{GPUs} & \textbf{Storage (5 iters)} & \textbf{Train Data} & \textbf{SPIDER Test (EX)} \\
\midrule
SFT      & 58 min       & 2 & 15.5 GB     & 7,000              &83.7\% \\
SPIN     & 6 hr 15 min  & 4 & 75.0 GB       & 7,000              &79.0\% \\
SPFT-SQL & 1 hr 45 min  & 2 & 18.5 GB     & 7,000 real + 3,000 syn &\textbf{87.4\%} \\
\bottomrule
\end{tabular}
\caption{Resource Consumption and Performance Comparison (Qwen2.5 Coder-7B)}
\label{tab:resource_comparison}
\end{table*}

Table~\ref{tab:resource_comparison} summarizes the computational time, GPU usage, storage requirements, and final execution accuracy (EX). Compared to SFT, SPFT-SQL roughly doubles computational time per iteration, mainly due to two sequential fine-tuning stages: (1) the Verification-Based Iterative Fine-Tuning (VBI-FT), which simultaneously optimizes both the Text-to-SQL and SQL-to-Text models, and (2) the subsequent Self-Play Fine-Tuning stage. Despite this increased overhead, SPFT-SQL achieves a significant improvement of 3.7\% in execution accuracy compared to SFT, which justifies the additional computational cost.

In terms of GPU resources, our approach matches the GPU usage of standard SFT, utilizing only two GPUs. This efficiency arises because the SQL-to-Text and Text-to-SQL models are alternately fine-tuned, incurring no additional parallel GPU usage. Conversely, SPIN requires twice as many GPUs due to its fine-tuning strategy.

Regarding storage, SPFT-SQL increases storage usage by only approximately 19\% compared to SFT across five iterations. This modest increase results from our efficient strategy of storing only model weights rather than entire models. In contrast, SPIN's storage requirements are approximately five times greater than those of SFT, as it saves full models at every iteration.


Thus, despite moderate increases in computational time and storage, SPFT-SQL provides substantial accuracy gains over conventional approaches, offering a balanced trade-off suitable for practical research environments.

\subsection{Comparison of Generated SQLs from Different Methods}
\label{Appendix:generate_sql}

To better illustrate the improvements of our method over others, we selected two examples from the SPIDER and BIRD datasets, as shown in Tables \ref{tab:example1} and \ref{tab:example2}. In the first example, both the models fine-tuned with SFT and SPIN lost the ability to compute ratios, while our method was able to correctly calculate the formula and select the correct fields. This is attributed to our VBI-FT method, which generates high-quality training data during the iterative process, and the guidance provided during the Self-Play phase, helping the model select the correct fields and generate accurate SQL queries. In the second example, both SFT and SPFT achieved the correct results, while SPIN produced incorrect values, further demonstrating the issue with the SPIN self-play mechanism, where the model's historical outputs are treated as negative samples.

\begin{table*}[ht!]
\centering
\small
\begin{tabularx}{\textwidth}{@{}p{3cm}X@{}}
\toprule
\textbf{Stage} & \textbf{Details} \\
\midrule
\textbf{Data} & 
\textbf{Question}: What is the eligible free or reduced price meal rate for the top 5 schools in grades 1-12 with the highest free or reduced price meal count of the schools with the ownership code 66? \\
& \textbf{Gold SQL}: SELECT CAST(T1.`FRPM Count (K-12)` AS REAL) / T1.`Enrollment (K-12)` FROM frpm AS T1 INNER JOIN schools AS T2 ON T1.CDSCode = T2.CDSCode WHERE T2.SOC = 66 ORDER BY T1.`FRPM Count (K-12)` DESC LIMIT 5 \\
\midrule
\textbf{Original Model} & 
\textbf{Pred SQL}: SELECT frpm.CDSCode, frpm.School Name, frpm.FRPM Count (K-12), frpm.Enrollment (K-12), (frpm.FRPM Count (K-12) / frpm.Enrollment (K-12)) AS Eligible\_FRPM\_Rate FROM frpm JOIN schools ON frpm.CDSCode = schools.CDSCode WHERE schools.OwnershipCode = 66 ORDER BY frpm.FRPM Count (K-12) DESC LIMIT 5; \\
& \textbf{Result}: Wrong \\
\midrule
\textbf{SFT} & 
\textbf{Pred SQL}: SELECT T2.`Percent (\%) Eligible Free (K-12)`, T2.`Percent (\%) Eligible FRPM (K-12)` FROM schools AS T1 INNER JOIN frpm AS T2 ON T1.CDSCode = T2.CDSCode WHERE T1.SOC = 66 ORDER BY T2.`FRPM Count (K-12)` DESC LIMIT 5; \\
& \textbf{Result}: Wrong \\
\midrule
\textbf{SPIN} & 
\textbf{Pred SQL}: SELECT "Percent (\%) Eligible Free (K-12)", "Percent (\%) Eligible FRPM (K-12)" FROM frpm WHERE "2013-14 CALPADS Fall 1 Certification Status" = 66 ORDER BY "FRPM Count (K-12)" DESC LIMIT 5; \\
& \textbf{Result}: Wrong \\
\midrule
\textbf{SPFT-SQL} & 
\textbf{Pred SQL}: SELECT cast(frpm.`frpm count (k-12)` AS REAL) / frpm.`enrollment (k-12)` FROM frpm INNER JOIN schools ON frpm.cdscode = schools.cdscode WHERE schools.soc = 66 ORDER BY frpm.`frpm count (k-12)` DESC LIMIT 5; \\
& \textbf{Result}: Right \\
\bottomrule
\end{tabularx}
\caption{Comparison of Generated SQLs from Different Methods on the BIRD Dataset}
\label{tab:example1}
\end{table*}

\begin{table*}[ht!]
\centering
\small
\begin{tabularx}{\textwidth}{@{}p{3cm}X@{}}
\toprule
\textbf{Stage} & \textbf{Details} \\
\midrule
\textbf{Data} & 
\textbf{Question}: Count the number of searches made by buyers. \\
& \textbf{Gold SQL}: SELECT count(*) FROM Users AS T1 JOIN User\_Searches AS T2 ON T1.user\_id = T2.user\_id WHERE T1.is\_buyer = 1; \\
\midrule
\textbf{Original Model} & 
\textbf{Pred SQL}: SELECT COUNT(us.search\_seq) FROM Users u JOIN User\_Searches us ON u.user\_id = us.user\_id WHERE u.is\_buyer = 'Yes'; \\
& \textbf{Result}: Wrong \\
\midrule
\textbf{SFT} & 
\textbf{Pred SQL}: SELECT count(*) FROM users AS T1 JOIN user\_searches AS T2 ON T1.user\_id = T2.user\_id WHERE T1.is\_buyer = '1' \\
& \textbf{Result}: Right \\
\midrule
\textbf{SPIN} & 
\textbf{Pred SQL}: SELECT count(*) FROM users AS t1 JOIN user\_searches AS t2 ON t1.user\_id = t2.user\_id WHERE t1.is\_buyer = 't' \\
& \textbf{Result}: Wrong \\
\midrule
\textbf{SPFT-SQL} & 
\textbf{Pred SQL}: SELECT count(*) FROM users JOIN user\_searches ON users.user\_id = user\_searches.user\_id WHERE users.is\_buyer = 1; \\
& \textbf{Result}: Right \\
\bottomrule
\end{tabularx}
\caption{Comparison of Generated SQLs from Different Methods on the SPIDER Dataset}
\label{tab:example2}
\end{table*}

\end{document}